\documentclass[10pt,conference]{IEEEtran}

\usepackage{graphicx} 
\usepackage{booktabs, multirow} 
\usepackage{soul}
\usepackage[table]{xcolor} 
\usepackage{changepage,threeparttable} 
\usepackage{amsmath,amssymb,amsfonts}
\usepackage{algorithmic}
\usepackage{graphicx}
\usepackage{textcomp}
\usepackage{xcolor}
\usepackage{makecell}
\def\BibTeX{{\rm B\kern-.05em{\sc i\kern-.025em b}\kern-.08em
    T\kern-.1667em\lower.7ex\hbox{E}\kern-.125emX}}
\usepackage{epsfig}
\usepackage{caption}
\usepackage{subcaption}
\usepackage{calc}
\usepackage{amssymb}
\usepackage{amstext}
\usepackage{amsmath}
\usepackage{amsthm}
\usepackage{multicol}
\usepackage{adjustbox}
\usepackage{pslatex}
\usepackage{algorithm2e}
\usepackage{rotating}
\usepackage[bottom]{footmisc}
\usepackage{tabularx}
\usepackage{siunitx}
\usepackage{array}
\usepackage{tabularray}

\usepackage{cleveref}

\usepackage[acronym]{glossaries}

\newacronym{dl}{DL}{Deep Learning}
\newacronym{ae}{AE}{Autoencoder}
\newacronym{tcn}{TCN}{Temporal Convolutional Network}
\newacronym{ann}{ANN}{Artificial Neural Network}
\newacronym{rnn}{RNN}{Recurrent Neural Network}
\newacronym{lstm}{LSTM}{Long Short-Term Memory}
\newacronym{gru}{GRU}{Gated Recurrent Unit}
\newacronym{mse}{MSE}{Mean Squared Error}
\newacronym{vae}{VAE}{Variational Autoencoder}
\newacronym{ml}{ML}{Machine Learning}
\newacronym{mit}{MIT}{Massachusetts Institute of Technology}
\newacronym{ecg}{ECG}{Electrocardiogram}
\newacronym{elbo}{ELBO}{Evidence Lower Bound}
\newacronym{kl}{KL}{Kullback-Leibler}
\newacronym{tpe}{TPE}{Tree-structured Parzen Estimators}
\newacronym{if}{IF}{Isolation Forest}

\title{Unsupervised Anomaly Detection in Process-Complex Industrial Time Series: A Real-World Case Study}

\author{
\begin{tabular}{ccc}
\begin{tabular}{c}
1\textsuperscript{st} Sergej Krasnikov\textsuperscript{*} \\
\textit{Universität Augsburg} \\
Augsburg, Germany \\
krasnikov.research@gmail.com \\
0009-0008-1034-7290
\end{tabular}
&
\begin{tabular}{c}
2\textsuperscript{nd} Lukas Meitz\textsuperscript{*} \\
\textit{Technische Hochschule Augsburg} \\
Augsburg, Germany \\
Lukas.Meitz@tha.de
\end{tabular}
&
\begin{tabular}{c}
3\textsuperscript{rd} Samineh Bagheri \\
\textit{Independent Researcher} \\
Karlsruhe, Germany \\
email address or ORCID
\end{tabular}
\\
\noalign{\vskip 1.5em}
\begin{tabular}{c}
4\textsuperscript{th} Michael Heider \\
\textit{Universität Augsburg} \\
Augsburg, Germany \\
michael.heider@uni-a.de \\
0000-0003-3140-1993
\end{tabular}
&
\begin{tabular}{c}
5\textsuperscript{th} Thorsten Schöler \\
\textit{Technische Hochschule Augsburg} \\
Augsburg, Germany \\
email address or ORCID
\end{tabular}
&
\begin{tabular}{c}
6\textsuperscript{th} Jörg Hähner \\
\textit{Universität Augsburg} \\
Augsburg, Germany \\
joerg.haehner@uni-a.de \\
0000-0003-0107-264X
\end{tabular}
\\
\noalign{\vskip 1.5em}
\multicolumn{3}{c}{\textsuperscript{*}These authors contributed equally to this work.}
\end{tabular}
}

\begin{document}

\maketitle

\begin{abstract}
Industrial time-series data from real production environments exhibits substantially higher complexity than commonly used benchmark datasets, primarily due to heterogeneous, multi-stage operational processes. 
As a result, anomaly detection methods validated under simplified conditions often fail to generalize to industrial settings. 
This work presents an empirical study on a unique dataset collected from fully operational industrial machinery, explicitly capturing pronounced process-induced variability.

We evaluate which model classes are capable of capturing this complexity, starting with a classical Isolation Forest baseline and extending to multiple autoencoder architectures. 
Experimental results show that Isolation Forest is insufficient for modeling the non-periodic, multi-scale dynamics present in the data, whereas autoencoders consistently perform better. 
Among them, temporal convolutional autoencoders achieve the most robust performance, while recurrent and variational variants require more careful tuning.
\end{abstract}




\section{Introduction}

Industrial environments generate large volumes of multivariate time-series data from heterogeneous sensors monitoring machinery and its executed processes. 
These streams are often high-dimensional, non-periodic, and exhibit multi-scale temporal dynamics, reflecting complex operational sequences and noise-prone conditions. 
Detecting deviations from normal behavior in such settings is a key prerequisite for reliable Machine Health monitoring and Predictive Maintenance.

Classical anomaly detection methods such as Isolation Forest (IF) are attractive in industrial practice due to their simplicity, low computational cost, and label-free training. 
However, they typically rely on learning static feature distributions and struggle when normal behavior is defined by non-stationary, multi-stage processes with variable temporal structure. 
This raises the question of whether classical methods remain viable under pronounced process-induced complexity, or whether representation-learning models are required.

Our study is built upon a proprietary dataset, that has been collected from more than 100 connected devices of the same type.
This real-world dataset contains all of the complexity that naturally comes with operation and usage in production scenarios, which makes it a valuable resource for studying the effect of complexity on different model architectures.

To address this subject effectively, we evaluate the capability of a simpler model, the \gls{if}, to serve as a basis of comparison and implement six \gls{ae} configurations: three standard and three variational—implemented with \gls{tcn}, \gls{lstm}, and \gls{gru} architectures.
The models are trained and tested on our proprietary industrial dataset characterized by non-periodic, multi-scale process dynamics and sensor-rich recordings, reflecting typical challenges in machine monitoring environments.

In summary, this work makes the following contributions: 
\begin{enumerate}
    \item We systematically evaluate anomaly detection methods of varying complexity on industrial time-series.
    \item We present a controlled empirical study comparing autoencoder architectures---convolutional versus recurrent, deterministic versus variational---in their ability to model process-complex (see \ref{sec:process-complex}) industrial sequences, addressing a gap in the literature where such systematic comparisons remain rare.
    \item We show that architectural alignment with temporal structure is more critical than model complexity: convolutional networks outperform recurrent alternatives through better capture of multi-scale process dynamics in real-world industrial data.
\end{enumerate}

Our findings highlight architectural and representational trade-offs, with convolutional standard \glspl{ae} demonstrating superior performance on complex industrial data.
While the results are derived from a single industrial application, the observed performance differences are driven by structural properties---non-periodicity, variable phase ordering, and process-induced non-stationarity---that are common across many real-world industrial processes.

\begin{figure*}[t]
    \centering
    \includegraphics[width=\linewidth]{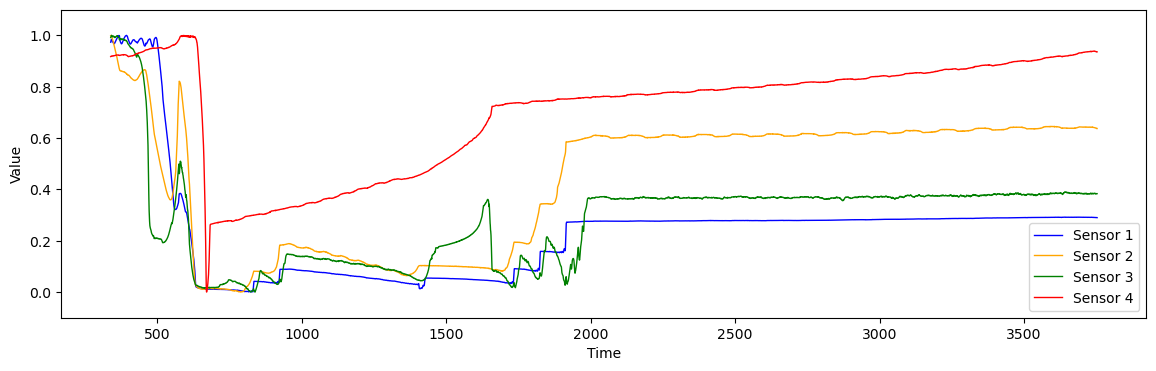}
    \caption{Plot of a process sample from the dataset, showing readings from four selected sensors captured during a single run executed on one machine. Each sensor records different physical quantities over time, exhibiting distinct behaviors including stable operational phases, gradual drifts, and abrupt changes indicative of process transitions or anomalies. The view of the data has been deliberately perturbed to protect sensitive operational information, while preserving the underlying structure and complexity.}
    \label{fig:multi_clean_example}
\end{figure*}


\section{Background}

Anomaly detection in time-series data has been extensively studied across domains such as healthcare, manufacturing, and infrastructure monitoring. In industrial settings, both classical and deep learning–based approaches are commonly employed, often under limited labeling and strict deployment constraints. However, a systematic understanding of how different model classes and architectures behave under realistic, process-induced complexity remains limited.

\subsection{Classical and Autoencoder-Based Anomaly Detection}

Classical anomaly detection methods, including IFs and related feature-based techniques, have long been used in industrial monitoring due to their simplicity, scalability, and low data requirements. 
These methods typically operate on static feature representations and implicitly assume stationarity, which limits their ability to model non-periodic, multi-stage process dynamics commonly observed in real production environments.

In recent years, deep learning approaches---particularly autoencoder-based models---have demonstrated their potential for unsupervised time-series anomaly detection. 
Autoencoders learn compact latent representations from unlabeled data and identify anomalies via reconstruction error~\cite{Rum1986}. 
Their effectiveness has been shown across a wide range of applications, including industrial systems and predictive maintenance. 
Recent surveys provide comprehensive overviews of deep anomaly detection methods and highlight the large design space of autoencoder architectures~\cite{SurveyDeepTSAD}.

A variety of autoencoder variants have been proposed to model temporal dependencies in time-series data. 
Recurrent architectures based on LSTM and GRU units are widely used for sequential modeling and have been applied to feature extraction and anomaly detection in industrial and energy systems~\cite{Yu2021Analysis}. 
Extensions incorporating memory mechanisms~\cite{Gao2023TSMAE} or attention-based designs, such as OmniAnomaly~\cite{OmniAnomaly}, aim to improve robustness in multivariate settings. 
In parallel, \glspl{tcn} leverage dilated convolutions to capture long-range dependencies with stable training behavior and have shown strong performance in healthcare, manufacturing, and general anomaly detection tasks~\cite{Thi2021,Asahi2021Process,Park2022TCAE,gopali2021comparative}.

\glspl{vae} extend standard autoencoders by learning probabilistic latent representations and have been explored for dimensionality reduction~\cite{Todo2022Dimension}, weakly supervised anomaly detection~\cite{Wu2024Weakly}, and data augmentation~\cite{Dodda2024Exploring}.
Despite their theoretical appeal, empirical evidence for consistent advantages of variational formulations over standard \glspl{ae} in industrial anomaly detection remains mixed.

Recent industrial-focused surveys further emphasize that many deep anomaly detection methods are still predominantly evaluated on simplified or benchmark-style datasets, limiting conclusions about their robustness under realistic operational variability~\cite{MEITZ2025111193}.

\subsection{Need for Complexity-Aware Evaluation}
\label{sec:process-complex}

Although a wide range of anomaly detection architectures has been proposed, few studies systematically compare classical and \gls{ae}-based models across convolutional, recurrent, and variational designs under realistic industrial conditions. 
Prior work has noted that general-purpose models often underperform when confronted with heterogeneous, non-stationary processes~\cite{Thi2021}.

Recent research has argued that such evaluation gaps are closely tied to unmodeled sources of machine- and process-induced complexity, motivating the use of explicit complexity taxonomies to reason about dataset characteristics and model assumptions~\cite{icinco24}. 
Following this taxonomy, we call our dataset process-complex, as it includes complexity predominantly induced by the process that is being monitored.
Addressing this gap is essential for deriving practical guidance on model selection for industrial anomaly detection beyond benchmark-oriented evaluations.


\begin{table*}[!ht]
\captionsetup{skip=10pt}
\centering
\setlength{\tabcolsep}{6pt}
\renewcommand{\arraystretch}{1.2}
\begin{tabular}{lll}
\toprule
\textbf{Model} & \textbf{Encoder Configuration} & \textbf{Decoder Configuration} \\
\midrule
TCN-AE   & Conv1D + TCN blocks + Avg Pooling & Inverse TCN blocks + Conv1D \\
TCN-VAE  & Conv1D + TCN blocks + Avg Pooling + Normal ($\mu$, $\sigma$) & Inverse TCN blocks + Conv1D \\
LSTM-AE  & 2 LSTM layers & 2 LSTM layers \\
LSTM-VAE & 2 LSTM layers + Normal ($\mu$, $\sigma$) & 2 LSTM layers \\
GRU-AE   & 2 GRU layers & 2 GRU layers \\
GRU-VAE  & 2 GRU layers + Normal ($\mu$, $\sigma$) & 2 GRU layers \\
\bottomrule
\end{tabular}
\caption{Architectural configurations of the \gls{ae} models used in the study. \gls{vae} models include an encoded normal distribution signified by mean ($\mu$) and standard deviation ($\sigma$).}
\label{tab:model_architectures}
\end{table*}

\section{Method}

In order to contribute to the research gap identified in the previous section, we will implement and compare multiple proven models in the anomaly detection domain.
This section presents the experimental methodology used to investigate the performance of different models and configurations for anomaly detection in complex industrial time-series data. 

The evaluation follows a two-stage process.
First, models are assessed for their reconstruction performance using process data captured from industrial machines.
Second, top-performing models are applied to an independent and labeled dataset from the same type of machinery containing real anomalies, enabling quantitative evaluation of their anomaly detection capabilities.
Using this process, our goal is to evaluate the anomaly detection capability of different models on a dataset with high process-complexity, reflecting a realistic industrial application.

\subsection{Description of the Used Dataset}

The dataset used in this study originates from an industrial process involving automated product creation. 
It consists of multivariate time-series recordings collected from 118 field-deployed machines, encompassing 334 individual process instances. 
Each instance captures the full temporal evolution of a manufacturing cycle, including signals from a heterogeneous set of sensors and actuators.
An illustrative example of a multivariate process instance is shown in Figure \ref{fig:multi_clean_example}, highlighting the variability and heterogeneity across sensor signals during a typical production cycle.

This dataset exhibits several characteristics that distinguish it from commonly used benchmarks in time-series modeling.
Most notably, it features numerous sensor readings, non-periodic behavior, and multi-scale process dynamics.
These attributes arise from the complex structure of the underlying process, which comprises multiple sequential and interleaved operations with variable durations and response patterns. 
The data is further affected by sensor inaccuracies and inter-device variability, introducing realistic noise patterns typical of industrial deployments.

In contrast to widely used benchmark datasets such as NASA Turbofan or bearing test rigs, which typically exhibit repetitive cycles or monotonic degradation patterns, the recorded processes show variable phase orderings, durations, and actuator interactions, violating common assumptions of temporal alignment and stationarity.

Following the complexity taxonomy proposed by \cite{icinco24}, the dataset qualifies as process-complex due to the concurrent presence of actuator diversity, multi-phase control logic, heterogeneous timing profiles, and non-repetitive process executions.
These properties make the dataset particularly challenging for classical anomaly detection methods that rely on static feature distributions, while favoring models capable of learning hierarchical and temporal representations.

To ensure data quality, only newly commissioned machines were selected and the recordings were limited to the first two months of operation. 
This sampling strategy reduces the likelihood of degraded behavior contaminating the training data, allowing the models to focus on learning normal operational patterns. 
Preprocessing steps included normalization of all sensor channels, selective downsampling to reduce redundancy, and segmentation using a sliding window to preserve local temporal dependencies.

\subsection{Model Architectures and Training}
\label{sec:architectures}

To assess which model types can capture process-induced complexity in industrial time-series data, we evaluate a classical anomaly detection baseline alongside a set of representation-learning models based on \glspl{ae}. 
This enables a direct comparison between feature-agnostic methods and sequence-aware architectures under identical data conditions.

\paragraph{Baseline Model}

Isolation Forest was selected as a representative classical baseline due to its widespread use in industrial monitoring and scalability to high-dimensional data, and ability to operate without labeled anomalies. 
The method isolates samples via random feature partitioning and is effective for detecting deviations in static feature distributions. 
Each process instance was represented using aggregated statistical features over the full sequence, following standard practice for classical anomaly detection on time-series data. 
This baseline serves to assess whether feature-agnostic methods remain viable for process-driven industrial data with complex temporal dynamics.

\paragraph{Autoencoder Architectures}
As primary models, we implemented six \gls{ae} variants: standard and variational based on \gls{tcn}, \gls{lstm}, and \gls{gru}. 
These architectures cover both convolutional and recurrent design paradigms commonly used in time-series modeling. 
Encoder and decoder configurations are summarized in Table~\ref{tab:model_architectures}.

All \gls{ae} models were implemented in PyTorch and trained using a unified protocol. 
Hyperparameter optimization was conducted using Optuna, with standard \glspl{ae} minimizing reconstruction error (MSE) and \glspl{vae} maximizing the \gls{elbo}. 
Training was performed on a single-node Databricks cluster with an NVIDIA T4 GPU.

Models were trained exclusively on data from normal machine operation to reflect realistic industrial conditions. 
Each variant was trained 50 times until convergence with different random seeds and hyperparameter configurations. 
The top five configurations per model, selected based on validation performance, were retained for downstream anomaly detection. 
The evaluated hyperparameter options are reported in Table~\ref{table:hyperparameters}.

\begin{table}[h!]
\centering
\renewcommand{\arraystretch}{1.3}
\begin{tabular}{lll}
\hline
 \textbf{Hyperparameter Description}        & \textbf{Values}            \\ 
 \hline
Gradient Descent Learning Rate              & (0.001, 0.0001)              \\
Batch Size                                  & \{64, 128, 256, 512\}           \\
Input Sequence Length                       & \{480, 600, 720, 840, 960, 1080\}\\
Hidden Layer Size                           & \{32, 64\}                      \\
Number of encoder/decoder Layers            & (1, 3)                        \\
Latent Dimension (VAE)                      & \{16, 32, 64\}                  \\
Kernel Size (TCN)                           & (2, 10)                     \\
Downsampling Factor (TCN)                   & \{4, 6, 8, 10\}                 \\ \hline
\end{tabular}
\caption{Hyperparameter search space for model training. Values represent ranges (min, max) for continuous parameters and discrete sets \{...\} for categorical parameters. Architecture-specific parameters are shown for \gls{vae} and \gls{tcn} models.}
\label{table:hyperparameters}
\end{table}

\subsection{Model Evaluation on Anomaly Detection}

To assess the practical utility of the trained models, we evaluate their anomaly detection capabilities using a separate, labeled dataset that was not involved in training or hyperparameter optimization. 
This evaluation dataset consists of 46 complete process instances, 22 of which were manually labeled as anomalous and 24 as normal. 
The dataset exhibits similar structure and sensor diversity as the training data but includes known deviations introduced through real operational failures or irregular process conditions.

Anomaly detection was performed by computing reconstruction error for each process instance.
A decision threshold was then applied to the aggregated error to classify each instance as anomalous or normal. 
Since reconstruction errors vary across models and configurations, the threshold was optimized individually based on the separate dataset for each model to balance detection performance.

To this end, a multi-objective optimization was applied using NSGA-II to find Pareto-optimal threshold values across four standard classification metrics: precision, recall, accuracy, and F1-score.
These metrics were selected to provide a comprehensive evaluation of model performance, particularly in the context of imbalanced classification tasks such as anomaly detection \cite[p.~412]{goodfellow2016deep}.

Detection thresholds were optimized post hoc for each model to estimate best achievable anomaly detection performance and do not represent fixed deployment thresholds.
The best-performing threshold for each model was retained to represent its potential performance in a real-world deployment scenario. 
This evaluation setup enables a direct comparison of models in terms of both their reconstruction-based learning capabilities and their practical effectiveness in detecting anomalous machine behavior.


\section{Results}

Following the experimental methodology described in the previous section, multiple models of increasing complexity were trained and evaluated on the dataset. 
Table~\ref{tab:model_performance_comparison} summarizes the average performance across all approaches.

\begin{table}[h!]
\captionsetup{skip=10pt}
\setlength{\tabcolsep}{4pt}
\renewcommand{\arraystretch}{1.3}
\centering
\begin{tabular}{lccccc}
\toprule
& \multicolumn{2}{c}{\textbf{Anomaly Detection}} & \multicolumn{2}{c}{\textbf{Reconstruction}} & \\
\cmidrule{2-3} \cmidrule{4-5}
\textbf{Model} & \textbf{F1 Score} & & \textbf{MSE / ELBO ($\times 10^{-4}$)} & \\
\midrule
\textbf{\gls{if}} & 0.120 $\pm$ 0.126 & & - & \\
\midrule
\textbf{TCN-AE}           & \textbf{0.991 $\pm$ 0.009} & & \textbf{0.22 $\pm$ 0.06} & \\
LSTM-AE                   & 0.853 $\pm$ 0.102 & & 1.23 $\pm$ 0.25 & \\
GRU-AE                    & 0.918 $\pm$ 0.066 & & 0.84 $\pm$ 0.15 & \\
\midrule
TCN-VAE                   & \textbf{0.968 $\pm$ 0.010} & & -7.29 $\pm$ 0.47 & \\
LSTM-VAE                  & 0.945 $\pm$ 0.027 & & \textbf{-1.89 $\pm$ 0.16} & \\
GRU-VAE                   & 0.876 $\pm$ 0.085 & & -2.24 $\pm$ 0.07 & \\
\bottomrule
\end{tabular}
\caption{Average performance metrics for the evaluated models in the presented industrial case study. Reported values are means $\pm$ standard deviation (MSE for standard AE, ELBO for VAE).}
\label{tab:model_performance_comparison}
\end{table}

\subsection{Isolation Forest (Baseline)}
\label{sec:iforest_results}

The results reveal that \gls{if} is fundamentally unsuited for this industrial anomaly detection task. 
Across all runs, the approach achieved an average F1-score of $0.120 \pm 0.126$. 
Performance varied dramatically between runs: five runs detected no anomalies whatsoever, while the best performing run achieved only an F1-Score of $0.308$ by detecting four out of 22 anomalous processes. 
Although the models achieved a precision of 1.0 in the runs in which anomalies were detected, the maximum recall of 0.182 indicates systematic failure to identify the majority of anomalous instances.

\subsection{Autoencoder Models}
\label{sec:ae_results}

\paragraph{Reconstruction Performance}

All \gls{ae} architectures learned to replicate the multivariate sensor sequences, demonstrating their fundamental suitability for this complex industrial dataset. 
However, reconstruction fidelity varied significantly between architectures.

The TCN-AE consistently achieved the lowest reconstruction error, outperforming both recurrent alternatives by a substantial margin. 
Beyond superior reconstruction quality, the convolutional model exhibited the highest consistency across training runs, underscoring its robustness to hyperparameter variations. 
GRU-AE achieved moderate reconstruction quality, while LSTM-AE showed both higher reconstruction error and greater variability across runs, indicating less stable convergence behavior.

\paragraph{Anomaly Detection Performance}

\begin{figure}[!h]
	\centering
	\includegraphics[width=\linewidth]{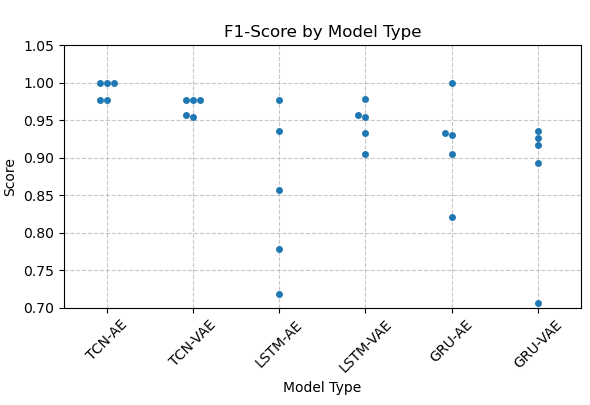}
	\caption{Differences in anomaly detection performance across models with TCN variants achieving the most desirable results. Shown are the F1-scores of the top 5 models for each architecture. Note: \gls{if} is excluded from this visualization due to its significantly lower performance (cf. \Cref{tab:model_performance_comparison}), which would distort the scale and clarity of the plot.}
	\label{fig:distribution}
\end{figure}

The primary evaluation criterion is practical applicability for anomaly detection. 
The results demonstrate a substantial improvement over the IF across all architectures. 
TCN-AE achieved most robust performance in this setting with some models even achieving an F1-Score of 1, detecting all anomalies correctly. 
GRU-AE followed with strong but more variable results, while LSTM-AE showed the highest variability and was more prone to both false positives and negatives. 
Figure~\ref{fig:distribution} illustrates these performance differences through F1-score distributions across the top 5 configurations for each architecture, for both standard and variational variants.

\subsection{Variational Autoencoder}
\label{sec:vae_results}

VAEs consistently underperformed their deterministic counterparts in anomaly detection. 
While LSTM-VAE and TCN-VAE achieved respectable performance, they fell short of the corresponding standard \glspl{ae}. 
GRU-VAE showed the most pronounced degradation.

TCN-VAE in particular showed occasional degraded reconstructions marked by noise artifacts or spikes, reflected in its high ELBO variance.
Among variational variants, LSTM-VAE achieved the best ELBO and most stable training behavior, though this did not translate to superior anomaly detection performance compared to the TCN models.

Training time analysis over variational and standard AEs revealed practical trade-offs: GRU-based models converged fastest (15–30 minutes), while LSTM and TCN variants required 40–60 minutes per model. 
However, given the scales at which industrial applications operate, these differences are likely negligible in practice, as inference performance is the more critical factor for deployment.

\section{Discussion}

This section interprets the experimental results with respect to the two central research questions, focusing on how architectural inductive biases affect anomaly detection performance under pronounced process-induced complexity. 
The reported results reflect best-case detection performance under controlled evaluation and should be interpreted as an assessment of model capacity rather than as a finalized deployment configuration.
Rather than reiterating quantitative results, the discussion emphasizes explanatory factors and practical implications for model selection in industrial monitoring scenarios.

While the presented results are derived from a single industrial application, the observed performance differences are driven by structural properties---non-periodicity, variable phase ordering, and process-induced non-stationarity---that are common across many real-world industrial processes.

\subsection{Architectural Effects under Process-Induced Complexity}

The results indicate a clear advantage of convolutional \gls{ae} architectures, particularly \gls{tcn}-based models, when applied to complex industrial time-series data. Their superior performance can be attributed to the ability of temporal convolutions to capture local and mid-range temporal dependencies while remaining robust to variable phase lengths and non-repetitive process structures. This inductive bias is well aligned with the characteristics of the studied dataset, which exhibits heterogeneous, multi-stage operational behavior rather than periodic or stationary dynamics.

Recurrent architectures based on \gls{lstm} and \gls{gru} were decent at modeling the data but showed increased sensitivity to hyperparameter choices and training conditions. Their sequential processing nature makes them more susceptible to instabilities when process stages vary in duration or ordering, which limits robustness in settings where extensive tuning is impractical.

Across all architectures, standard \glspl{ae} consistently outperformed their variational counterparts. While \glspl{vae} provide theoretical advantages through regularized latent spaces, the introduced stochasticity proved detrimental in a reconstruction-driven anomaly detection setting. In particular, increased variance and reconstruction artifacts reduced threshold stability, indicating that precise deterministic reconstruction is more critical than latent expressiveness for detecting subtle process deviations.

Taken together, these observations suggest that architectural robustness and alignment with process structure are more decisive than model complexity alone. For process-complex industrial time-series, convolutional \glspl{ae} offer a favorable balance between representational power, stability, and practical deployability.

\subsection{Limitations and Outlook}

This study is limited to a single proprietary industrial dataset, which restricts direct comparison with publicly available benchmarks. Consequently, the findings should be interpreted as qualitative guidance on architectural suitability under process-induced complexity rather than as universal performance rankings. In addition, the evaluated model set was restricted to classical baselines and \gls{ae}-based architectures.

While attention-based architectures such as transformer-based \glspl{ae} are promising for modeling state-rich or highly variable processes, their data requirements, tuning complexity, and computational cost often conflict with the limited fault data and resource constraints typical of industrial anomaly detection deployments.


\section{Conclusion}

This work evaluated anomaly detection models on a real-world industrial time-series dataset characterized by pronounced process-induced complexity, including non-periodic behavior, heterogeneous sensors, and multi-stage operational dynamics. 
Such characteristics are largely absent from commonly used benchmark datasets and pose a substantial challenge for anomaly detection methods.

By comparing a classical \gls{if} baseline with multiple \gls{ae} architectures, the study demonstrates that feature-agnostic classical methods are insufficient for reliably modeling complex industrial processes. 
\gls{ae}-based models consistently achieved superior performance, with architectural choice playing a decisive role. 
In particular, \gls{tcn}-\glspl{ae} provided the most robust and stable results, while recurrent and variational variants required more careful tuning and showed higher performance variability.

These findings underline, within the studied industrial application, the necessity of representation-learning approaches for anomaly detection in process-complex industrial time-series and highlight temporal convolution as a particularly effective design choice. 
Overall, the results provide practical guidance for model selection in industrial monitoring scenarios where realistic process variability must be addressed.

\bibliographystyle{IEEEtran}
\bibliography{references.bib}

\end{document}